%% file: main.tex
\documentclass[journal,twocolumn]{IEEEtran}

\IEEEoverridecommandlockouts                              
\usepackage{cite}
\usepackage{subcaption}
\usepackage{graphicx}
\usepackage{amsmath}
\usepackage{amssymb}
\usepackage[normalem]{ulem}
\usepackage[ruled,vlined]{algorithm2e}
\usepackage{framed}
\usepackage[table]{xcolor}
\usepackage{array}
\usepackage{booktabs, tabularx}
\usepackage{multirow}
\usepackage{bigints}
\usepackage{times}

\usepackage[]{algorithm2e}
\usepackage{bm}
\usepackage{color}
\usepackage{ulem}
\graphicspath{ {images/} }

\newcommand{\prob}{\mathbb{P}}
\usepackage{balance}

\newcommand{\mv}[1]{{\color{black}#1}}
\newcommand{\jd}[1]{{\color{black}#1}}
\newcommand{\jdp}[1]{{\color{black}#1}}

\newcommand{\yudel}[1]{}
\usepackage{float}
\usepackage{caption}

\title{Air-Aided Communication \\ Between Ground Assets in a Poisson Forest \vspace{0.01cm}}
\author{
\IEEEauthorblockN{Juan David Pabon, Shaikha Alkandari, Matthew C. Valenti, and Xi Yu
 } \\
West Virginia University, Morgantown, WV, USA. \\
\vspace{-0.25cm}
}

\begin{document}

\maketitle
\thispagestyle{empty}
\pagestyle{empty}

%%%%%%%%%%%%%%%%%%%%%%%%%%%%%%%%%%%%%%%%%%%%%%%%%%%%%%%%%%%%%%%%%%%%%%%%%%%%%%%%
\begin{abstract}
\input{sections/0_abstract.tex}

\end{abstract}

\vspace{-0.2cm}

\section{Introduction}
\label{intro}
\input{sections/1_introduction.tex}

\section{Problem formulation}
\label{ProbForm}

\input{sections/2_section.tex}

\section{Probability of Obtaining LoS}
%\subsection{Formation of the communication links}
\label{section3}

\input{sections/3_section.tex}
\section{Calculating the Throughput}
\label{section4}

\input{sections/4_section.tex}

\section{Simulation-Validated Numerical Results}
\label{simulation}
\input{sections/5_simulation.tex}

\section{Conclusions and future work}
\label{conclusion}
\input{sections/6_conclusion.tex}

% \section{Appendix}
% \label{appendix}
% \input{sections/7_appendix.tex}


%%%%%%%%%%%%%%%%%%%%%%%%%%%%%%%%%%%%%%%%%%%%%%%%%%%%%%%%%%%%%%%%%%%%%%%%%%%%%%%%

\bibliographystyle{IEEEtran}
\balance
\bibliography{library}

\end{document}

%% file: sections/0_abstract.tex
Ground assets deployed in a cluttered environment with randomized obstacles (\textit{e.g.}, a forest) may experience line of sight (LoS) obstruction due to those obstacles. Air assets can be deployed in the vicinity to aid the communication by establishing two-hop paths between the ground assets. Obstacles that are taller than a position-dependent critical height may still obstruct the LoS between a ground asset and an air asset. In this paper, we provide an analytical framework for computing the probability of obtaining a LoS path in a Poisson forest.   Given the locations and heights of a ground asset and an air asset, we establish the critical height, which is a function of distance.   To account for this dependence on distance, the blocking is modeled as an inhomogenous Poisson point process, and the LoS probability is its void probability.  Examples and closed-form expressions are provided for two obstruction height distributions: uniform and truncated Gaussian.  The examples are validated through simulation.  Additionally, the end-to-end throughput is determined and shown to be a metric that balances communication distance with the impact of LoS blockage.  Throughput is used to determine the range at which it is better to relay communications through the air asset, and, when the air asset is deployed, its optimal height.  
%Results for an example system operating at 38 GHz show that the air asset should be used when the communication range is greater than about 50 meters.

% We then calculated the location-dependent critical height for any obstacle according to its distance from the ground asset and the air asset, as well as the height at which the air asset flies, and modeled the distribution of the obstacles that may block the line of sight as an Inhomogeneous Poisson point process. Numerical examples of calculating the probabilities of connections in different scenarios are provided, as well as conditions under which air assets should be introduced in aiding the connections between ground assets. The results are validated via simulations. 

%% file: sections/1_introduction.tex
% Paragraph I - 

% \textit{Todo: define the purpose of this paper, which could be}
% \begin{itemize}
%     \item determine the connection probability between robots in forest (done)
%     \item determine the conditions that UAV aided communication outperforms the direct communication between ground robots (done)
%     \item optimal location for a UAV (maybe future work)
%     \item optimal deployment of multiple UAVs (maybe future work)
% \end{itemize}

Many military communication scenarios involve the deployment of heterogeneous teams of mobile assets that execute joint tasks in challenging environments, such as forests,  where randomized obstacles are located.  Often these tasks are autonomous and the assets may be robotic. Potential application scenarios range from exploring and monitoring sensitive areas to search and rescue \cite{langelaan2005towards,zheng2010multirobot,harikumar2018multi,couceiro2019semfire,tian2020search,oliveira2021advances}. In such scenarios, different mobile assets in the same team need to coordinate with each other and perform tasks jointly. 
Information sharing is an essential piece to the coordination of the team members \cite{doriya2015brief} and can be realized via data streaming from one asset to another.
%or one asset monitoring and modeling the environments including other team members. mcv: I wasn't sure what was meant at the end of that sentence.

% A myriad of modern communication techniques (\textit{e.g.}, mmWave) have been advanced to achieve efficient data streaming. 

Computer-aided localization and mapping \cite{mysorewala2009multi,benjamin2015real} relying on cameras and laser devices have been developed to realize high-quality environment modeling in forests. 
Camera or laser enabled modeling and monitoring techniques usually require the mobile asset to continuously see the objects (\textit{e.g.}, its teammates) being monitored.  Cutting-edge communication technologies use short wavelengths, such as in millimeter-wave (mmwave) or even terahertz (THz) bands, to enhance the transmission speed, but the penetration capability is limited and usually requires a line of sight (LoS) to close a link. Motivated by these examples and others like it, it is clear that maintaining a line of sight (LoS) is preferred between these kinds of mobile assets.

%Communication in environments with obstacles is particularly challenging due to loss of LoS, shadowing, signal decaying, etc. 

Various existing works related to environments with obstacles focus on planning for the best routes for mobile assets or the optimal locations for stationary assets to avoid the connections between certain pairs of assets being blocked by the obstacles \cite{gasparetto2015path}. Such methods rely on perfect knowledge of the locations of the obstacles. In reality, this type of knowledge may be expensive to acquire, especially for large-scale areas. 

Locations of obstacles randomly distributed in a cluttered 2-D environment can be seen as generated via a Poisson point process (PPP). Such a model is usually referred to as a \textit{Poisson Forest}. Extensive work has been done related to navigating mobile assets in such forests \cite{karaman2012high,karaman2012high2,junior2021fast}. Probabilities of having LoS of a minimum length in such a 2-D environment are also computed \cite{karaman2012high} and related to the density of the PPP.  

Because a given team may deploy heterogeneous mobile assets located at different altitudes (\textit{e.g.} both ground assets and air assets), it is essential to consider the 3-D nature of the environment.  Not only the thickness and the locations of the obstacles need to be considered, but also the \emph{heights} of the obstacles are important. References \cite{hriba2017accurately,hriba2018impact,hriba2021optimization} address the height distribution of obstacles in an urban environment with semi-randomly distributed obstacles. References \cite{baccelli2015correlated,Gapeyenko} model urban areas with obstacles of random heights using the Manhattan Point Line Process to address the grid-like patterns of the obstacles' locations. This series of work points out that only obstacles taller than a \textit{critical height} can block the LoS between a ground asset and an air asset. The critical height is a location-dependent function that depends on the ground and the air assets' locations as well as the height that the air asset is flying at.   Any obstacle that is taller than the critical height evaluated at that obstacle's location will block the LoS.   

In this paper, we study the 3-D LoS between heterogeneous assets in a Poisson forest. We calculate the critical height, i.e., the height above which an obstacle is able to block the LoS, at any given location between a pair of assets. The distribution of the obstacles above the critical height is then modeled as an \textit{inhomogeneous} PPP. By finding the void probability of the inhomogeneous PPP, we are able to determine the probability of obtaining LoS. 
Examples and expressions that are in closed form (up to known standard functions) are provided for two obstruction height distributions: uniform and truncated Gaussian.  The examples are validated through simulation.  Additionally, we establish \emph{throughput} as a metric that balances communication distance with the impact of LoS blockage.  Throughput is used to determine the range at which it is better to relay communications through the air asset, and, when the air asset is deployed, its optimal height.

The paper is organized as follows. Sec.~\ref{ProbForm} formulates the problem. Sec.~\ref{section3} analyzes the probability of obtaining LoS between a pair of assets. Sec.~\ref{section4} discusses the throughput of the communication among assets. Sec.~\ref{simulation} provides numerical results derived from our methods as well as simulation validations. Sec.~\ref{conclusion} concludes the paper.

%Paragraph II - Popular robots applications in challenging (with obstacles) environments such as forest. Multi-robot applications. Communication (\textit{Todo: find a better expression that includes communication and camera vision sights}) is essential. Communication is challenging when there are obstacles due to shadowing, signal decay, etc. (\textit{Todo: cite papers})

%Paragraph III - Modern communication techniques require unobstructed vision - borrow the LoS concept. Various work planned for the best routes/locations in such environments to avoid being affected by the obstacles. (\textit{Todo: cite trajectory planning papers}) Such methods rely on known location of the obstacles. Not the randomness of the obstacles. 

%Paragraph IV - Existing works addressing 2-D randomness of obstacles (cite Frazzoli 2012 CDC paper) like Poisson Forests. While heterogeneous teams of robots are considered (UGV and UAV), 3-D models need to be considered - height is important.(\textit{check papers that cited Frazzoli's paper})

%Paragraph V - Existing works addressing obstacles heights - (cite Valenti Milcom paper using the Manhattan model). This model is for grid setting. This paper points out that the critical heights blocking LoS depends on the locations of the obstacles and the robots. 

%Paragraph VI - This paper takes a Poisson forest, and generates an in-homogeneous Poisson point process model for the trees that are `tall enough'. We considered both the UGV-UGV case and the UGV-UAV case. Providing conditions that one can outperform the other. 

%% file: sections/2_section.tex
\label{problemFormulation}
Consider a pair of ground assets, each equipped with a communication device (\textit{e.g.}, an antenna, a camera, etc.) at a height of $h_g$, deployed in a planar task space (\textit{e.g.}, a forest) with stochastically distributed obstacles (\textit{e.g.}, trees) of a non-trivial thickness. The locations of the obstacles are generated by a two-dimensional Poisson Point Process with a fixed density $\lambda_f$. Let $N$ be the number of obstacles in a task space of area $A_f$ and $\lambda_f$ be the intensity of the PPP representing the expected number of obstacles per unit area.  From the basic properties of a PPP
\begin{align*}
    \prob \{N=n\} = \frac{({\lambda_f A_f)}^n}{n!}e^{-\lambda_f A_f},
\end{align*}
Such a task space is referred to as a \textit{Poisson forest}.

\jd{
The height of any single obstacle in the Poisson forest is represented by a non-negative random variable $H_t$. The distribution of $H_t$ may vary. We denote the cumulative distribution function (cdf) of $H_t$ as $F_{H}(h)$. Evaluating $F_{H}(h)$ at a given height $h_0$ gives the probability that a given obstacle has a height that is less than or equal to $h_0$. 

The analysis in this paper does not require that $H_t$ assumes any particular distribution. In fact, the height distribution of trees in a forest varies \cite{kohyama1989frequency,felfili1997diameter,mauro2011influence}. Bell-shaped distributions \cite{felfili1997diameter} and positive-skewed distributions \cite{kohyama1989frequency} can both be found. 
To provide specific realistic cases, without any loss of generality, we consider a single-sided truncated Gaussian distribution and a uniform distribution as examples to illustrate our methods.  Both examples are constructed such that the heights are non-negative.  

The truncated Gaussian uses a Gaussian random variable with mean $\mu$ and standard deviation $\sigma$ as its parent distrution, and is truncated to the range $h \geq 0$.  The cdf of this variable is
\begin{align}\label{densityBuildings}
        F_{H}(h)= \dfrac{Q\left(\dfrac{\mu -h}{\sigma} \right) - Q\left(\dfrac{\mu}{\sigma} \right)}{1 - Q\left(\dfrac{\mu}{\sigma} \right)},
\end{align}
for $h \geq 0$, and zero elsewhere, where $Q(\cdot)$ is the Q-function.

For the uniform distribution, the cdf is 
\begin{equation}\label{densityUniform}
     F_{H}(h)= \left\{
     \begin{array}{ll}
             0 & \quad  \text{for } h < 0 \\
            \dfrac{h}{h_{\max}} & \quad  \text{for }  0 \leq h \leq h_{\max} \\
            1  & \quad  \text{for } h > h_{\max} 
        \end{array}
   \right.,
\end{equation}
where $h_{\max}$ is the maximum height of the obstruction. 

Now consider the one-dimensional space between the two ground assets. Let the locations of the two ground assets be $0$ and $x_g$ on this 1-D coordinate system. Any obstacle with a height above $h_g$ located along the interval $(0,x_g)$ would potentially block the unobstructed view of one ground asset on the other. If there is no such obstacle, we consider there is a \textit{line of sight} (LoS) between the two ground assets.

In the above-mentioned Poisson forest, the expected number of obstacles located exactly along the one-dimensional space $(0,x_g)$ should be zero since the Lebesgue Measure of a straight line in a two-dimensional space is always zero. However, when the obstacles' thicknesses are non-trivial, obstacles located in a finite area with a non-trivial width around the line $(0,x_g)$ may also block the LoS between the two ground assets. Therefore we consider the distribution of potential obstacles along a straight line in this Poisson forest to be characterized by a 1-D Poisson Process with a fixed density $\lambda_0$ capturing the expected number of obstacles located along a straight line of unit length. $\lambda_0$ is determined by $\lambda_0 = E(w) \lambda_f$, where $E(w)$ is
the average thickness of obstacles.

Now consider the case that an air asset at an altitude or height of $h_a$ is available to aid in the communication, for instance, by receiving the signal transmitted to it by the first ground asset and relaying it to the second ground asset.   In this paper, we limit the horizontal location of the air asset to be along the straight line $(0,x_g)$ that connects the two ground assets.  If we define $x_a$ to be the horizontal location of the air asset, then $0 \leq x_a \leq x_g$.    

In the following sections, we calculate the probabilities of obtaining a LoS in ground-ground, ground-air, and ground-air-ground (i.e., air-aided) connections, and calculate the throughput in all these scenarios. We furthermore find the throughput for these cases.  Our calculations provide conditions under which air assets should be deployed to aid the connections between ground assets and provide insight into the optimal height of the air asset when one is deployed.

}

%% file: sections/3_section.tex
In this section, we provide methods for calculating the probabilities of obtaining a LoS (\textit{i.e.}, the \emph{LoS probability}), between a pair of ground assets as well as between one ground asset and one air asset. 

\begin{figure}[t]
\centering
\includegraphics[width=0.45\textwidth]{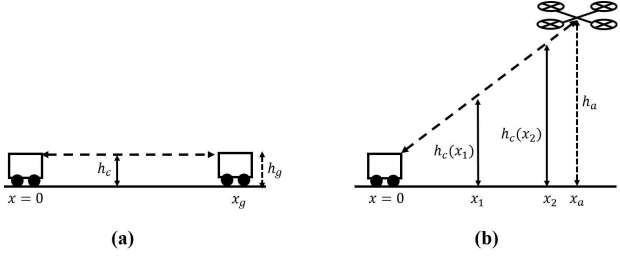}
\caption{(a) shows the critical height of obstacles between a pair of ground assets. (b) shows the critical height determined by the straight line between a ground asset and an air asset.}
\label{fig:CritHeight}
\end{figure}

\jd{
\subsection{Critical height}\label{sec:critical}
In the Poisson forest introduced in the previous section, only obstacles that are above a certain height will block the view between two assets. If the two assets are both on the ground and have their communication devices at the same height $h_g$, only obstacles with heights greater than $h_g$ may block the LoS. In this case, we say $h_g$ is the critical height $h_{c}$ of the obstacles. 

When considering the LoS between a ground asset of height $h_g$ and an air asset of height $h_{a}$, the critical height is determined by the straight line connecting the communication devices of the ground asset and the air asset which is located at $x_a$, as shown in Fig.~\ref{fig:CritHeight}}.
\jd{Given both assets are fixed, the critical height is a function of the location  $x \in [0, x_{a}]$ along the horizontal coordinate.  As can be seen in Fig.~\ref{fig:CritHeight}}, the critical height is low close to the ground asset, where even short obstructions can block the LoS, but is high further away from the ground asset, where only the tallest obstructions can block the LoS.  The critical-height function is 
\begin{equation}
    h_{c}(x)=\dfrac{h_{a}-h_{g}}{x_{a}}x + h_{g}.
    \label{hcritical}
\end{equation}
Notice that $h_{c}(x)$ increases linearly with $x$ increasing from $0$ to $x_{a}$.

\subsection{Inhomogeneous Poisson Point Process}\label{sec:InhomoPPP}
As introduced in Sec.~\ref{ProbForm}, the location of obstacles are generated by a homogeneous PPP with density $\lambda_0$. Consider the case that the critical height $h_c$ is a constant, such as with the direct ground-ground link.  Since not all the obstacles are tall enough to block the LoS between assets, we must retain only those obstacles whose heights are higher than $h_{c}$. For any random obstacle, the probability that it has a height $h_o$ greater than $h_{c}$ can be calculated by 
\begin{align*}
    \prob\{h_o>h_{c}\} = 1-F_{H}(h_{c}).
\end{align*}
%where $F_{H}(\cdot)$ is the cumulative distribution function of the obstacles' height. 
Therefore, the distribution of obstacles with heights greater than $h_{c}$ in this Poisson forest can be modeled as a Poisson process with its density $\lambda_{c}$ defined as 
\begin{equation}
    \lambda_{c}=\lambda _{0} \prob\{h_o>h_{c}\} = \lambda _{0}\left[ 1-F_{H}\left(h_{c}\right) \right].
    \label{lambdavv}
\end{equation}

When $h_{c}$ is location-dependent as in \eqref{hcritical}, this model becomes an inhomogeneous Poisson process with a location-dependent density $\lambda(x)$ defined as
\begin{align}
\label{eq:InhomoDensity}
    \lambda(x) = \lambda _{0}\left[1-F_{H}\left(h_{c} \left(x \right)\right) \right].
\end{align}

For a given $F_{H}(\cdot)$ (e.g \eqref{densityBuildings} or \eqref{densityUniform}), the cumulative probability towards a given $h$ increases with $h$. Therefore, $\lambda(x)$ decreases monotonically when $h$ increases. That is to say, in a given Poisson forest, with a greater critical height, the probability of having obstacles taller than this critical height is lower.

\subsection{LoS between a pair of ground assets}
As discussed in Sec.~\ref{sec:critical},  only obstacles with heights above $h_{g}$ will block the direct view between a pair of ground assets. Therefore we set the critical height $ h_{c}=h_{g}$. Since $h_{c}$ is constant, the blockages are described by  the homogeneous PPP given by \eqref{lambdavv}. 
%The distribution of obstacles along the straight line $(0,x_j)$ that are tall enough to block the LoS can be modeled as 
%\begin{align*}
%    \prob \{N_{gg}=n\} = \frac{\left({\lambda _{0} \prob\{h_o>h_{cr}\}}\right)^n}{n!}\exp\left(-\lambda_0 \prob\{h_o>h_{cr}\}\right),
%\end{align*}
%where $N_{gg}$ is the number of obstacles located on a unit length of straight line between these ground assets. 
Therefore, the probability of obtaining a LoS between a pair of ground assets that are located at $0$ and $x_g$ is found from the void probability of the homogeneous PPP of density $\lambda_c = \lambda _{0}\left[ 1-F_{H}\left(h_{c}\right) \right]$ over an interval of length $x_g$. This probability is
%\begin{align}
%    \prob_{LoS,gg}(d_{gg}) = e^{-\lambda_0\prob\{h_o>h_{c}\} d_{gg} }.
%\label{Plosvv}
%\end{align}
\begin{align}
    \prob_{LoS}^{gg}(x) = e^{-\lambda _{0}\left[ 1-F_{H}(h_{g}) \right]x_g },
\label{Plosvv}
\end{align}

\subsection{LoS between a ground asset and an air asset}
The distribution of obstacles along the straight line from the ground asset ($x=0$) to the air asset ($x=x_{a}$) can be modeled as a Poisson process with a density of $\lambda_0$. The critical height, above which an obstacle may block the LoS between the two assets, varies depending on the location of the obstacle. Therefore, the distribution of the obstacles that are tall enough to disrupt the unobstructed view fits an inhomogeneous Poisson process with a location-dependent density defined as in \eqref{eq:InhomoDensity}.

%\jd{ The distribution is as follows
%\begin{align}
%    \prob_{LoS}(x_{i}) = \exp\left( -\dfrac{1}{x_{i}} \int_{0}^{x_{i}} \lambda(x_{i},x)x_{i} dx \right),
%    \label{Plos}
%\end{align}
%where $x \in [0, x_{i}]$. }

%In Fig.~\ref{lambdaXFig}, we can see that as the distance from some point in the horizontal axis to the vehicle increase, the value of $\lambda(x_{i},x)$ decreases. It is because for some determined $x_{i}$, as $x$ increases, the critical height increases and the amount of buildings that could be taller than the critical height is reduced.

%Therefore, the density of the Poisson Point Process changes as function of the critical height (which changes depending on $x_{i}$ and $x$). Figure \ref{lambdaXFig} shows an example of how $\lambda(x_{i},$x$)$ changes.  
% As discussed in Sec.~\ref{sec:InhomoPPP}, $\lambda(x)$ is monotonically decreasing with $x$.
%when $x$ moves from the ground asset at $x=0$ to the air asset at $x=x_{a}$. In this case, 
The probability of having no obstacle blocking the LoS between a ground asset at $0$ and an air asset at $x_{a}$ is found from the void probability of the inhomogeneous PPP, which is found as follows
\begin{equation}
    \prob_{LoS}^{ga}(x) = \exp \left(- \int_{0}^{x_{a}} \lambda(x) dx \right).
    \label{Plos}
\end{equation}
Finding a solution to ~\eqref{Plos} depends on the difficulty of performing the integration, which depends on the nature of $F_{H}(h)$.  In some cases, such as the two examples provided below, closed-form solutions can be found, at least up to being expressed in terms of well-known expressions.  Alternatively, when closed-form solutions cannot be readily obtained,  ~\eqref{Plos} can be solved numerically, for instance, by using numerical integration.

%The probability of having no obstacles taller than the critical height \textit{within a unit length} at a given $x \in [0, x_{0}]$ is 
%\begin{equation}
%    \prob_{LoS}(x_{i}, x) = e^{-\lambda(x_{i},x)x_{i}}.
%    \label{EQPLoSxix}
%\end{equation}
%$\prob_{LoS}(x)$ increases with $x$ for $x_{0}$ given. 

% It is not always trivial to find analytical solutions to eq.~\eqref{Plos} with different types of $F_{H}(h)$. However, eq.~\eqref{Plos} shows that the probability of a pair of assets obtaining LoS is determined by the horizontal distance in between. Computational tools can be easily found to determine the exact value of $\prob_{LoS}^{ga}(x)$ whenever needed. 

For the two different heights distributions mentioned in Sec.\ref{problemFormulation}, we can find the analytical solutions of \eqref{Plos}. Sec. \ref{analyticalGaussian} and Sec.\ref{analyticalUniform} show the analytical solutions for the truncated Gaussian and uniform distributions.

\subsection{$P_{LoS}^{ga}$ for truncated Gaussian distribution}

\label{analyticalGaussian}
For the truncated Gaussian distribution, the density of the inhomogenous PPP is given by  \eqref{eq:InhomoDensity}, where $F_H(\cdot)$ is given by \eqref{densityBuildings} and $h_c(\cdot)$ is given in \eqref{hcritical}.
%as follows. In this case $\lambda(x)$ is given by eq. \eqref{eq:InhomoDensity}, where $h_c(x)$ is determined by the critical height in eq. \eqref{hcritical}. Substituting eq. \eqref{densityBuildings} into $\lambda(x)$ we get
%{\small \begin{equation*}
%\lambda(x) = \lambda_{0}\left( \dfrac{1 - %Q\left(\dfrac{ \mu - h_{c}(x)}{\sigma}\right)}{1 %- Q\left( \dfrac{\mu}{\sigma}\right)}\right),
%\end{equation*}
%}
%where $Q(\cdot)$ is the Q-function. 
%From the properties of the Q-function, we have $1-Q(\cdot) = \Phi(\cdot)$, where $\Phi(\cdot)$ is the cumulative distribution function. We then obtain
The resulting density is
{\small\begin{equation*}
\lambda(x) = \lambda_{0}\left( \dfrac{\Phi\left(\dfrac{ \mu - h_{c}(x)}{\sigma}\right)}{\Phi\left( \dfrac{\mu}{\sigma}\right)}\right).
\end{equation*}
}
where 
\begin{eqnarray}
\Phi(z) & = & 1 - Q(z) = 
\dfrac{1}{2}\left(1 + \text{erf}\dfrac{z}{\sqrt{2}}\right)
\label{phi}
\end{eqnarray}
is the cdf of the standard normal distribution. 

The integral in \eqref{Plos} is
{\small\begin{equation*}
    \int_{0}^{x_{a}}\lambda(x)dx = \dfrac{\lambda_{0}}{{\small \Phi\left(\dfrac{\mu}{\sigma}\right)}}\bigintss_{0}^{x_{a}}\Phi\left( \dfrac{\mu - h_{c}(x)}{\sigma}\right)dx.
\end{equation*}
}
From (\ref{phi}), this can be rewritten as
\begin{eqnarray}
    \int_{0}^{x_{a}}\lambda(x)dx 
     =
    c\int_{0}^{x_{a}}\left( 1 + \text{erf}(a-bx) \right) dx \nonumber \\
     = 
    c\left(x_a +  \int_{0}^{x_{a}}\text{erf}(a-bx)dx \right) 
    \label{LambdaGaussian}
\end{eqnarray}
where $a=(\mu - h_{g})/(\sqrt{2}\sigma)$, $b=(h_a -h_{g})(\sqrt{2}\sigma x_{a})$ and $c= \lambda_{0}/\left(2\Phi\left(\mu/\sigma\right)\right)$.  

% pagebreak needed because the column was getting compressed for some reason.
\pagebreak

The integral on the second line of \eqref{LambdaGaussian} can be found from mathematical handbooks (e.g. \cite{gradshteyn2014table}) to be
{\footnotesize
\begin{equation}
\dfrac{e^{-a^{2}}+\sqrt{\pi}\left[ (bx_{a}-a)\text{erf}(a-bx_{a})+a\text{erf}(a) \right] - e^{-(a-bx_{a})^{2}}}{\sqrt{\pi}b}.
\label{lastintegral}
\end{equation}
}
The LoS probability is then found by substituting \eqref{LambdaGaussian} into \eqref{Plos}
with the integral in the second line of \eqref{LambdaGaussian} set to \eqref{lastintegral}.

% This result provides the probability of obtaining LoS between a ground and an air asset at a fixed height when the horizontal location of the air asset is $x=x_{a}$, and the obstacles' heights distribution fits a truncated Gaussian distribution.% is considered.

\jdp{

\subsection{Analytical solution of $P_{LoS}^{ga}$ for uniform distribution}
\label{analyticalUniform}
For the uniform distribution, the density of the inhomogenous PPP is given by  \eqref{eq:InhomoDensity}, where $F_H(\cdot)$ is given by \eqref{densityUniform} and $h_c(\cdot)$ is again given in \eqref{hcritical}.
Evaluating $F_{H}(h_{c}(x))$ we obtain
\begin{equation}
     F_{H}(h_{c}(x))= \left\{
     \begin{array}{ll}
             \quad 0 & \quad  \text{for } x \leq x' \\
            \left(\dfrac{h_{a} - h_{g}}{x_{a}h_{\max}}\right)x +\dfrac{h_{g}}{h_{\max}} & \quad  \text{for }  x' < x \leq x_{c} \\
            \quad 1  & \quad  \text{for } x > x_{c} 
        \end{array}
   \right.
   \label{analyticalUniformCumulativeDistribution}
\end{equation}
where $x' = -h_{g}x_{a}/(h_{a} - h_{g})$ and $x_{c}$ is the critical distance  at which any obstacle located at a distance $x>x_{c}$ cannot block the LoS since $h_{c}(x) > h_{\max}$. This distance is given by
\begin{equation*}
    x_{c} = \left(\dfrac{h_{\max} - h_{g}}{h_{a} - h_{g}}\right)x_{a}
\end{equation*}
since $x'$ is negative it is inconsequential, since the integral of \eqref{Plos} does not cover negative $x$.  
The density of the inhomogeneous PPP is found by substituting \eqref{analyticalUniformCumulativeDistribution} into \eqref{eq:InhomoDensity} resulting in
{\small
\begin{equation}
     \lambda(x)= \left\{
     \begin{array}{ll}
            \lambda_{0}\left[ 1 -\left(\dfrac{h_{a} - h_{g}}{x_{a}h_{\max}}\right)x -\dfrac{h_{g}}{h_{\max}}\right] & \text{for }  0 < x \leq x_{c} \\
            \quad 0  &   \text{for } x > x_{c}. 
        \end{array}
   \right.
   \label{lambdaUniform}
\end{equation}
}

From \eqref{Plos}, $\prob_{LoS}^{ga}(x_{a})$ requires that $\lambda(x)$ be integrated from 0 to $x_{a}$. If $x_{a}\leq x_{c}$, then 
\begin{equation}
    \int _{0}^{x_{a}} \lambda(x)dx = \int _{0}^{x_{a}} \lambda_{0}\left[ 1 -\left(\dfrac{h_{a} - h_{g}}{x_{a}h_{\max}}\right)x -\dfrac{h_{g}}{h_{\max}}\right] dx.
    \label{integralLambda}
\end{equation}
When $x_{a}>x_{c}$, $\lambda(x)=0$ for $x>x_{c}$ and thus the integral from $x_c$ to $x_a$ is zero.  It follows then that the integral will take the same form as in \eqref{integralLambda} but the upper limit can be tightened to $x_{c}$ since the integral beyond that point is zero.  

Rather than expressing the integral separately for the two cases of $x_{a}\leq x_{c}$ and $x_{a}>x_{c}$, we can express them as the following single expression
{\small
\begin{equation*}
    \int_{0}^{x_{a}}\lambda(x)dx = \lambda_{0}\int_{0}^{\min(x_{a}, x_{c})}\left[ 1 -\left(\dfrac{h_{a} - h_{g}}{x_{a}h_{\max}}\right)x -\dfrac{h_{g}}{h_{\max}}\right] dx. 
\end{equation*}
}
defining $\lambda_{u}$ as the solution of the previous integral we have 
{\small
\begin{equation*}
\lambda_{u} =\lambda_{0}\min(x_{a}, x_{c})\left[ 1 - \dfrac{h_{g}}{h_{\max}} - \left( \dfrac{h_{a -h_{g}}}{2x_{a}h_{\max}}\right)\min(x_{a}, x_{c}) \right].
\end{equation*}
}

% pagebreak needed because the column was getting compressed for some reason.
%\pagebreak

Then, using $\lambda_{u}$ we get that the probability of obtaining LoS for ground-air communication $\prob_{LoS}^{ga}$ at $x=x_{a}$ is given by
{\small
\begin{equation*}
 \prob_{LoS}^{ga}(x_{a})=e^{-\lambda_{u}}.
\end{equation*}
}

}

%% file: sections/4_section.tex
\mv{
While the LoS probability is useful for predicting the existence of a LoS between two assets, it does not characterize the \emph{quality} of the link, which is also affected by the transmission distance of each link.   For instance, if one wants to determine the height of an air asset that maximizes \textit{only} the end-to-end LoS probability with a ground asset, the solution would be to place the air asset at an infinite height so that the critical heights at any location between the two assets are infinite (\textit{i.e.}, no obstacle is able to block the unobstructed view between them).  Such solution is neither practical nor efficient in reality due to the significant signal loss caused by the infinite transmission distance. % However, an infinite height is  infeasible not only for practical reasons but also because there would be significant loss in signal power.  

An appropriate metric that captures the loss of signal power at distance is the expected \emph{throughput}, which we here define to be the maximum achievable data rate when accounting for the possibility of blockage.  This definition makes sense considering a mixed team of ground and air assets navigating through the forest and maintaining communication. The probability of LoS between any pair of assets yields an expected communication time, which contributes to an expected throughput that can be achieved.  For a single hop, the expected throughput is
\begin{equation}
   T = \prob_{LoS} C \label{throughput},
\end{equation}
where $C$ is the \emph{capacity} of the link, and the multiplication by $\prob_{LoS}$ accounts for the expectation being with respect to LoS.   
%This interpretation of throughput is consistent with the modern use of hybrid-ARQ protocols, where a blocked transmission would trigger a retransmission.  
Here, we set $C$ as the Shannon Capacity, which is the maximum achievable rate of an unblocked link
\begin{equation}
  C = B\log_{2} \left( 1+\mathsf{SNR} \right) \label{capacity},
\end{equation}
where $B$ is the signal bandwidth, and the signal-to-noise ratio, when expressed in dB, is
\begin{equation}
  \mathsf{SNR}^\mathsf{dB}
  =
  \mathsf{SNR}_0^\mathsf{dB}
  -10\alpha \log_{10}{\left(\frac{d}{d_{0}}\right)} \label{SNR},
\end{equation}
where $\alpha$ is the path-loss exponent, $d_0$ is a reference distance typically set to $1$ meter, and $\mathsf{SNR}_0^\mathsf{dB}$ is the SNR when the receiver is placed at distance $d_0$ assuming free-space propagation up to that distance.    The value of $\mathsf{SNR}_0^\mathsf{dB}$ can be measured, or it can be calculated from the transmit power, carrier frequency, bandwidth, receiver's noise figure, and antenna gains.

For a direct ground-ground transmission, the expected throughput is computed from \eqref{throughput} with $d$ in \eqref{SNR} set to $x_g$.   For a two-hop ground-air-ground transmission, it depends on the LoS probabilities of both hops.   Let $\prob_{LoS}^\mathsf{ga}$ and $\prob_{LoS}^\mathsf{ag}$ be the LoS probabilities of the ground-to-air and air-to-ground links, respectively, and similarly define $C_\mathsf{ga}$ and $C_\mathsf{ag}$ as the two capacities, the expected throughput for the ground-air-ground communication is
\begin{equation}
   T = \frac{1}{2} \prob_{LoS}^\mathsf{ga}  \prob_{LoS}^\mathsf{ag} \min(C_\mathsf{ga},C_\mathsf{ag}),
\end{equation}
where the multiplication by $1/2$ accounts for the time-division duplexing (TDD) operation at the air asset (\textit{i.e.}, the air asset spends half its time receiving from the first ground asset and half its time transmitting to the second ground asset).  
Alternatively, frequency-division duplexing (FDD) can be used, but in that case, the per-link capacities should be scaled accordingly since only half of the band could be used for each hop.  Each capacity is found from \eqref{capacity} with the distance in \eqref{SNR} set as the Euclidean distance between the air antenna and the corresponding ground antenna, where each distance is the hypotenuse of a right triangle formed with one leg being the horizontal distance, either $x_a$ for the ground-air link or $x_g-x_a$ for the air-ground link, and the other leg being the difference in antenna heights, $h_a-h_g$.  
Notice that the ground-air-ground throughput is determined by the minimum capacity of the two hops, which motivates us to consider deploying the aiding air asset always above the midpoint of the two ground assets.

}

%% file: sections/5_simulation.tex
%If we suppose that the height of the buildings is a random variable, we can use the normal distribution to analyze the behavior of the height of the buildings. Let $\mu$ be the mean of the height and $\sigma$ its standard deviation, then the probability density function of the building's height is given by

%\begin{equation}
%    f(h)= \dfrac{1}{\sigma \sqrt{2\pi}} e^{ \dfrac{1}{2}\left(\dfrac{h-\mu}{\sigma}\right)^2}.
%    \label{densityBuildings}
%\end{equation}

 %Choosing some values for the previous equation, we obtain a plot as the shown in Figure \ref{PdensityFunction}, where we can see the relation of the height of the buildings and its probability density function. Integrating \eqref{densityBuildings} from zero to some height $h$, give us the probability that a building has a height less than  or equal to $h$.   

%\begin{figure}[ht]
%\centering
%\includegraphics[width=0.45\textwidth]{Fig1.eps}
%\caption{Probability density function of the height $h$ of the buildings for $\mu = 19$ and $\sigma = 10$.}
%\label{PdensityFunction}
%\end{figure}
%\subsection{Numerical results}
In this section, we present numerical results generated by our methods. The key part of these methods is the calculation of the probability of obtaining a LoS between different types of assets. We consider a Poisson forest where the locations of the obstacles along a straight line are generated via a Poisson point process with $\lambda_0=0.02$. The distributions of the obstacle height $H_t$ were chosen to be a truncated Gaussian distribution with cdf $F_{H}(h)$ defined as in \eqref{densityBuildings} with $\mu = 19$ m and $\sigma = 10$ m and a uniform distribution with cdf $F_{H}(h)$ defined as in \eqref{densityUniform} with $h_{\max}=29$ m. The choice of the parameters is consistent with the parameters chosen in \cite{Gapeyenko}.

\mv{To validate the numerical results, we performed Monte Carlo simulations involving the repeated drawing of Poisson forests.   Let $\ell$ be the interval of a simulated Poisson forest, where $\ell=x_g$ for a ground-ground link or $\ell=x_a$ for a ground-air link.   Drawing a Poisson forest involves first determining the random number $N$ of obstructions in the interval, which is done by drawing $N$ from a Poisson distribution of mean $\lambda_0 \ell$.  Next, each of the $N$ obstructions is placed uniformly over the interval.  Then, each obstruction's height is determined by randomly drawing its $H$ from the corresponding distribution. Once the forest was constructed, it was determined whether or not the LoS path was blocked by checking to see if any of the obstructions were above the critical height at that location.  This process was repeated for 500 thousand trials for each data point reported.}

%\subsection{$P_{LoS}$ Vehicle - UAV }
%To study the behavior of $P_{LoS}(x_{i})$ some numerical simulation are performed. Since in the real world the only parameters that can be changed are the height of the UAV and the height of the vehicle antenna, we will only change these parameters and fix $\sigma$, $\mu$, and $\lambda_{0}$, since they are determined by the work environment.

%According with the information presented in Table III and Figure 6.b in \cite{Gapeyenko}, for the simulation We consider $\mu=19$, $\sigma=10$, and $\lambda_{0}=0.02$.       

%\begin{figure}[ht]
%\centering
%\includegraphics[width=0.45\textwidth]{Fig2.eps}
%\caption{Probability of having LoS at a distance $x_{i}$ from the vehicle for different values of $h_{uav}$. The units of $x_{i}$, $h_{uav}$, and $h_{v}$ are meters.}
%\label{fig2}
%\end{figure}
\jd{
Fig.~\ref{fig4} shows the probability of obtaining LoS between a ground asset and an air asset, $\prob_{LoS}^{ga}(x)$, with the horizontal distance $x_{a}$.  In this figure, only the truncated Gaussian distribution is considered, as results for the uniform distribution are similar. The air asset flies at different fixed heights of $50$, $100$, and $200$ meters. The ground asset has a communication device fixed at the height of $h_g = 2$ meters.
\begin{figure}[t]
\centering
\includegraphics[width=0.45\textwidth]{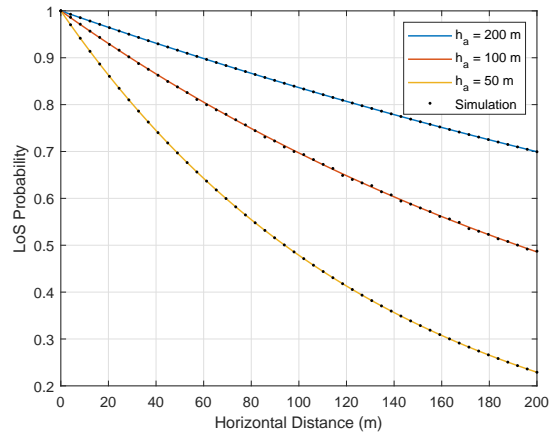}
\caption{Probability of obtaining LoS from a ground asset located at $0$ and to an air asset located at $x_{a}$ for a truncated Gaussian height distribution and $h_{g}=2$ m. Solid lines show the numerical results calculated using our closed-form expressions, while the dots show results generated by Monte Carlo simulation.}
\label{fig4}
\end{figure}
The results in Fig.~\ref{fig4} show that $\prob_{LoS}^{ga}(x)$ decreases as $x_a$ increases. The reason is straightforward. A longer distance between the two assets simply allows a greater probability of having obstacles in between. Meanwhile, $\prob_{LoS}^{ga}(x)$ increases as $h_{a}$ increases. This is because the critical height will increase with a greater $h_{a}$ (as in \eqref{hcritical}). A greater critical height rejects more obstacles from potentially blocking the LoS between the two assets. Therefore flying the air asset at a higher altitude generally increases the probability of obtaining LoS.

% slope of the straight line determined by \eqref{hcritical} (critical height) is greater for greater UAV heights. Taller buildings will have less probability of blocking the line of sight when the height of the UAV increases.  

%Fig.~\ref{fig3} shows the scenarios in which the air asset flies at a fixed height of $50$ meters, and the ground asset has a communication device fixed at different heights of $0$, $5$, $10$, and $15$ meters. The results are compared with flying an air asset at $100$ meters while fixing the ground asset's communication device at a height of $0$.

%\begin{figure}[ht]
%\centering
%\includegraphics[width=0.35\textwidth]{Fig3.eps}
%\caption{Probability of having LoS between a ground asset and an air asset with a distance $x_{i}$ in between for different $h_{uav}$ and $h_{v}$. The units of $x_{i}$, $h_{uav}$, and $h_{v}$ are meters.}
%\label{fig3}
%\end{figure}

Increasing the height $h_{g}$ of the communication devices carried by the ground asset will improve the probability of obtaining LoS as well. Generally, increasing the height of the ground assets' communication devices is a more expensive and less efficient way to enhance the probability of LoS as compared with increasing the height of the air asset. For practical purposes, a big increase of $h_{g}$ is not preferable, but a small increase may result in an acceptable increase of the $\prob_{LoS}^{ga}(x)$ for shorter distances.

We then compare the end-to-end LoS probability $\prob_{LoS}$ of direct ground-ground communication with air-aided ground-air-ground communication. When making this comparison, the air asset is always deployed above the midpoint of the two ground assets; \textit{i.e.}, $x_a = x_g/2$. In this scenario, the probability to obtain LoS from the air asset to both ground assets synchronously is the square of $\prob_{LoS}^{ga}$. We assume that the air asset is flying at a height of $h_a = 100$ m, while all ground assets have their communication devices fixed at a height of $h_g = 2$ m. Both the truncated Gaussian and the uniform height distributions are considered. Fig. \ref{comparison} shows the results of this comparison. For direct ground-ground communication, the probability of obtaining LoS decreases much faster as a function of distance than in the case of air-aided ground-air-ground communication. For the truncated Gaussian distribution, when $\mu$ = $19$ m and $\sigma$ = $10$ m, \eqref{densityBuildings} suggests that most of the obstacles will be taller than $2$ m. Thus, almost all obstacles can block the unobstructed view between a pair of ground assets, severely decreasing the probability of obtaining the LoS. For the uniform distribution, according to \eqref{densityUniform}, there is a probability greater than $0.92$ that the heights of the obstacles are taller than $h_{g}$. This causes a fast decrease in the $\prob_{LoS}$ for the ground-ground communication, which is  similar to what is observed for the truncated Gaussian distribution. 

 When $x_{a}=60$ m, (\textit{i.e.} $x_{g} = 120$ m), the probability of obtaining LoS between ground assets using direct ground-ground communication is approximately 0.1 for both the truncated Gaussian and the uniform distributions. However, when an air asset is used, the probability that it obtains a LoS with both ground assets is approximately $6.5$ and $7.3$ times greater than the probability of the two ground assets obtaining LoS over a direct link considering the truncated Gaussian and uniform distributions, respectively. Fig. \ref{comparison} shows that the choice of distribution does not have a significant impact on the probability of obtaining the LoS between ground assets using the direct link, since the communication devices of the ground assets are fixed at a relatively low height and therefore the LoS would be easily blocked by most obstacles. On the other hand, when an air asset is used, the height distribution has a bigger impact on the LoS probability since the differences of the distributions become more pronounced.

\begin{figure}[t]
\centering
\includegraphics[width=0.43\textwidth]{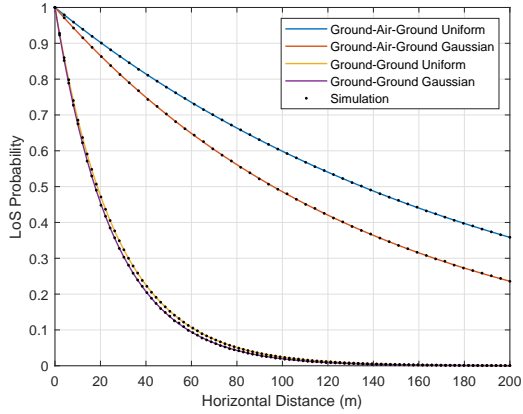}
\caption{Probabilities of obtaining the LoS for ground-air-ground and ground-ground communication for $h_a = 100$ m and $h_g = 2$ m. Solid lines show the numerical results calculated using our closed-form expressions, while the dots show results generated by Monte Carlo simulation.}
\label{comparison}
\end{figure}

}

% throughput results go here.
\mv{
In addition, we computed the throughput performance for the same scenarios previously discussed.  The additional parameters required to compute the throughput are a reference SNR of $\mathsf{SNR}_0^\mathsf{dB} = 51.98$ dB at a reference distance of $d_0 = 1$ meter, a path-loss coefficient of $\alpha = 2.3$, and a bandwidth of $B=20$ MHz.  This path-loss coefficient corresponds to the one reported in \cite{Rappaport2013} for the measured LoS pathloss at $38$ GHz.  The reference SNR is computed for a transmit power of $0$ dBm, a receiver noise figure of $9$ dB, and antenna gains of $12.1$ dBi for both the transmit and receive antennas, which are the gains reported for a compact 6-element array operating at $38$ GHz in \cite{Rahayu2018}.  We consider the same obstacle models as before, with $\lambda = 0.02$ and height distributions that are either a truncated Gaussian (with $\mu = 19$ and $\sigma = 10$) or a uniform (with $h_{max} = 29$).  The ground asset's antenna height is set to $h_g = 2$ m.

\begin{figure}[t]
\centering
\includegraphics[width=0.43\textwidth]{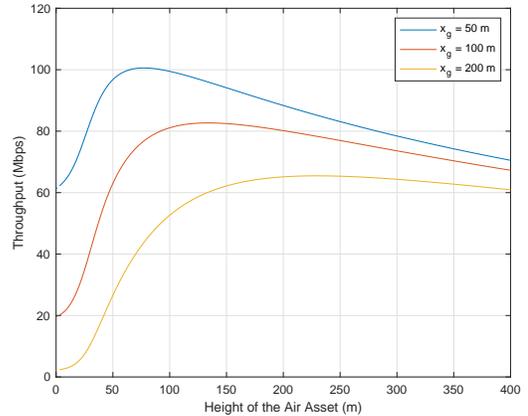}
\caption{Throughput of ground-air-ground communication as a function of the height $h_a$ of the air asset for horizontal distances $x_g = \{50,100,200\}$ m considering a Gaussian truncated distribution and $h_{g}=2$ m. }
\label{throughputfig1}
\end{figure}

Fig.~\ref{throughputfig1} shows throughput as a function of the height of the air asset, $h_a$, for several different distances between ground assets, $x_{g}$.   The air asset is located at the midpoint between the two ground assets, \textit{i.e.} $x_a = x_g/2$, and this figure shows results for just the truncated Gaussian height distribution (results for the uniform distribution are similar).  As expected, the throughput is higher when the ground assets are closer to each other.   However, for each curve, a peak value can be observed.  Lowering the altitude of the air asset below this peak makes it prone to blocking, but raising it above the peak value causes a loss in signal power which translates to a loss of capacity.   The peak value balances the assets' capability of obtaining LoS and the signal power, which is a key tradeoff as both contribute to the throughput.  For $x_{g}$ equal to $50$ m, $100$ m, and $200$ m, the peak values are $100.6$ Mbps, $82.7$ Mbps, and $65.5$ Mbps, respectively, and these peaks occur at $h_a$ of $77$ m, $134$ m, and $230$ m, respectively.

\begin{figure}[t]
\centering
\includegraphics[width=0.43\textwidth]{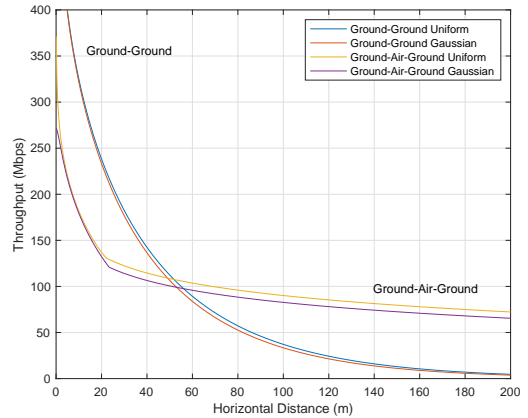}
\caption{Throughput as a function of horizontal distance $x_g$ for both truncated Gaussian and uniform height distributions. Results are shown for direct ground-ground communication as well as for relayed ground-air-ground communication.  In the case of ground-air-ground communication, the throughput is optimized with respect to the height $h_a$ of the air asset. }
\label{throughputfig2}
\end{figure}

Fig.~\ref{throughputfig2} shows throughput as a function of the horizontal distance $x_g$ between the ground assets.  The figure shows results for both truncated Gaussian and uniform height distributions and for both direct ground-ground communication and relayed ground-air-ground communication.  For ground-air-ground communication,  the throughput is optimized at each distance by maximizing its value over the height of the air asset $h_a$.   For direct ground-ground communication, no such optimization is possible.  The plot shows that, for sufficiently far distances, the throughput of the ground-air-ground communication is higher than that of the direct ground-ground communication.  However, for shorter distances, ground-ground communication has a higher throughput.  When the height distribution is a truncated Gaussian, this crossover occurs at a distance of $x_{g} = 52.9$ m, where the throughput for both direct ground-ground and relayed ground-air-ground communications is $99$ Mbps.  The reason that direct ground-ground communications performs better at ranges closer than this crossover distance is primarily due to the need for the air asset to duplex the signal received from the first ground asset and transmitted to the second ground asset.  The direct link does not need to duplex. However, at longer distances, maintaining a direct link between the two ground assets suffers from a lower probability of obtaining a LoS and a weaker signal power due to the long single transmission path.

}

%% file: sections/6_conclusion.tex
%With the results observed in simulations, we can conclude that if the model of the probability of the service quality is improved, given some work environment (represented by the Poisson Point Process), we can determine the distance from which the best option for achieving a mmWave communication between two vehicles is deploying an UAV and implementing a two-hop communication. 

%\textit{Todo: Future work}

In this paper, we studied issues related to deploying an air asset in a Poisson forest to aid the connection between a pair of ground assets. The key contribution is a framework for calculating the LoS probability and the throughput.   This framework depends on carefully considering the location-dependent critical height, which is the minimum height required for an obstruction at that location to block the LoS.  Because the critical height is distance dependent, the distribution of obstacles that are above the critical height is an inhomogeneous Poisson point process even if the location of the obstructions themselves is a homogeneous PPP.   Closed-form results are provided for two height distributions: truncated Gaussian and uniform.  Simulation results validate the theoretical expressions.  

The theory enables the solution to two particular problems. First, it allows the determination of a crossover distance, below which it is better for the ground assets to communicate directly, and above which air-assistance is desirable.  Second, it allows for the determination of the optimal height of the air asset when one is used.  The key to solving these problems is to use throughput as the performance metric, as throughput can properly balance the tradeoff wherein higher altitudes increase the LoS probability but reduce signal power. 

% such that random obstacles with a greater height may block the LoS between a pair of ground assets or between one ground asset and an air asset, and modeled the distribution of obstacles as an Inhomogeneous Poisson point process. The probabilities of connections in different scenarios are calculated numerically and validated via simulations. \jdp{With this, the throughput of the connections can be determined to find the optimal height to locate the air asset given a distance to the ground vehicles and determine the distances at which the throughput is higher using ground-ground or air aided communication.}     

This work is a gateway to optimally deploying a heterogeneous team of mobile air assets and ground assets in a dense forest. Future works include planning for the optimal deployment of the air assets given a known or unknown layout of the ground assets, and planning the optimal trajectories for the joint team to navigate through the task space while staying connected and coordinating on task delivery.